\documentclass[11pt,a4paper]{article}
\usepackage[T1]{fontenc}
\usepackage[utf8]{inputenc}
\usepackage[hyperref]{acl2018}
\usepackage{url}

\usepackage{times}

\usepackage{amsmath}
\usepackage{graphicx}
\usepackage{booktabs}
\usepackage{tabularx}
\usepackage{xcolor}
\usepackage{multirow}

\usepackage{tipa}

\usepackage{xspace}

\DeclareUnicodeCharacter{0167}{\texttstroke}
\DeclareUnicodeCharacter{0166}{\textTstroke}
\newcommand\texttstroke{t\llap{-\kern.07em}}
\newcommand\textTstroke{T\raise.1ex\llap{-\,}}

\newcommand{\task}[3]{{$\langle$#1, #2, #3$\rangle$\xspace}}

\definecolor{cb_green2}{RGB}{102,194,165}
\definecolor{cb_red}{RGB}{252,141,98}
\definecolor{cb_blue}{RGB}{141,160,203}
\definecolor{cb_purple}{RGB}{231,138,195}
\definecolor{cb_green1}{RGB}{166,216,84}
\definecolor{cb_yellow}{RGB}{255,217,47}
\definecolor{cb_salmon}{RGB}{229,196,148}
\definecolor{cb_gray}{RGB}{179,179,179}

\usepackage{tikz}
\usepackage{pgffor}
\usetikzlibrary{matrix,positioning,shapes.geometric, arrows, fit}
\usetikzlibrary{calc}
\tikzstyle{module_color} = [fill=cb_gray!70]
\tikzstyle{module} = [
rectangle, rounded corners, text centered,
module_color,
minimum height=.8cm, draw=black
]
\tikzstyle{large_module} = [
module,
text width=7cm,
minimum width=7cm,
]

\tikzstyle{bert_module} = [
module,
text width=9cm,
minimum width=10cm,
]

\tikzstyle{small_module} = [
module,
text width=1.5cm,
minimum width=1.5cm,
]

\tikzstyle{sequence} = [
minimum height=0.6cm,
rectangle, text centered,
draw=black, fill=cb_yellow!80,
]
\tikzstyle{target_sequence} = [
sequence, fill=cb_red!90,
]
\tikzstyle{output} = [
rectangle, text centered,
draw=black, fill=cb_green2!80
]

\tikzstyle{arrow} = [thick,->,>=stealth]
\tikzset{
	*|/.style={
		to path={
			(
			perpendicular cs:
			horizontal line through={(\tikztostart)},
			vertical line through={(\tikztotarget)}
			)
			-- (\tikztotarget) \tikztonodes
		}
	}
}

\tikzset{
	box/.style={rectangle,draw=black,thick, minimum size=0.01cm},
}

\def\rowdist{1.5}
\def\matrixheight{0.8}
\def\matrixwidth{0.4}

\newcommand*{\drawbertoutput}[2]{%
	\begin{scope}[shift={(#1-\matrixwidth/2, #2-\matrixheight/2)}]
		\foreach \x in {0,0.1,...,0.4}{
			\foreach \y in {0,0.1,...,0.8}{
				\draw [line width=0.05pt, fill=cb_yellow!75, draw=black] (\x, \y) rectangle (\x+0.1, \y+0.1);
			}%
		}%
	\end{scope}
}

\newcommand*{\drawberttargetoutput}[2]{%
	\begin{scope}[shift={(#1-\matrixwidth/2, #2-\matrixheight/2)}]
		\foreach \x in {0,0.1,...,0.4}{
			\foreach \y in {0,0.1,...,0.8}{
				\draw [fill=cb_red!90, line width=0.05pt, draw=black] (\x, \y) rectangle (\x+0.1, \y+0.1);
			}%
		}%
	\end{scope}
}

\aclfinalcopy 

\title{Evaluating Transferability of BERT Models on Uralic Languages}

\author{Judit Ács \\
    SZTAKI\\Institute for Computer Science\\and Control\\
  {\tt judit@sch.bme.hu} \\\And
  Dániel Lévai \\
  Department of Digital Humanities \\
  Eötvös Loránd University \\
  {\tt levai753@gmail.com} \\\And
  András Kornai \\
  SZTAKI\\Institute for Computer Science\\and Control\\
{\tt kornai@sztaki.hu} \\
}

\date{}

\begin{document}
\maketitle
\begin{abstract}

Transformer-based language models such as BERT have outperformed previous
models on a large number of English benchmarks, but their evaluation is often
limited to English or a small number of well-resourced languages.  In this
work, we evaluate monolingual, multilingual, and randomly initialized language
models from the BERT family on a variety of Uralic languages including
Estonian, Finnish, Hungarian, Erzya, Moksha, Karelian, Livvi, Komi Permyak,
Komi Zyrian, Northern Sámi, and Skolt Sámi. When monolingual models are
available (currently only et, fi, hu), these perform better on their native
language, but in general they transfer worse than multilingual models or
models of genetically unrelated languages that share the same character
set. Remarkably, straightforward transfer of high-resource models, even
without special efforts toward hyperparameter optimization, yields what appear
to be state of the art POS and NER tools for the minority Uralic languages
where there is sufficient data for finetuning.

A BERT- és más Transformer-alapú nyelvmodellek számos angol tesztadaton jobban
teljesítenek, mint a korábbi modellek, azonban ezek a tesztadatok az angolra
és néhány hasonlóan sok erőforrással rendelkező nyelvre korlátozódnak. Ebben
a cikkben egynyelvű, soknyelvű és random súlyokkal inicializált BERT
modelleket értékelünk ki a következő uráli nyelvekre: észt, finn, magyar,
erza, moksa, karjalai, livvi-karjalai, komi-permják, komi-zürjén, északi számi és
kolta számi. Az egynyelvű modellek -- jelenleg csak észt, finn és magyar
érhető el -- ugyan jobban teljesítenek az adott nyelvre, általában rosszabbul
transzferálhatóak, mint a soknyelvű modellek vagy a nem rokon, de azonos írást
használó egynyelvű modellek. Érdekes módon a sok erőforráson tanult modellek
még hiperparaméter optimalizálás nélkül is könnyen transzferálhatók és
finomhangolásra alkalmas tanítóadattal csúcsminőségű POS és NER taggerek
hozhatóak létre a kisebbségi uráli nyelvekre.

\end{abstract}

\section{Introduction}

Contextualized language models such as BERT \citep{Devlin:2018a} drastically
improved the state of the art for a multitude of natural language processing
applications. \citet{Devlin:2018a} originally released 4 English and 2
multilingual pretrained versions of BERT (mBERT for short) that support over
100 languages including three Uralic languages: Estonian [et], Finnish [fi],
and Hungarian [hu].  BERT was quickly followed by other large pretrained
Transformer \citep{Vaswani:2017} based models such as RoBERTa
\citep{Liu:2019a} and multilingual models such as XLM-RoBERTa
\citep{Conneau:2019a}. Huggingface released the Transformers library
\citep{Wolf:2020}, a PyTorch implementation of Transformer-based language
models along with a repository for pretrained models from community
contribution\footnote{\url{https://huggingface.co/models}}. This list now
contains over 1000 entries, many of which are domain-specific or monolingual
models.

Despite the wealth of multilingual and monolingual models, most
evaluation methods are limited to English, especially for the early models.
\citet{Devlin:2018a} showed that the original mBERT outperformed existing
models on the XNLI dataset \citep{Conneau:2018b}, a translation of the
MultiNLI \citep{Williams:2018b} to 15 languages.  mBERT was further evaluated
by \citet{Wu:2019} for 5 tasks in 39 languages, which they later expanded to
over 50 languages for part-of-speech (POS) tagging, named entity recognition
(NER) and dependency parsing \citep{Wu:2020a}.  mBERT has been applied to a
variety of multilingual tasks such as dependency \citep{Kondratyuk:2019} and
constituency parsing \citep{Kitaev:2019}. The surprisingly effective
multilinguality of mBERT was further explored by \citet{Dufter:2020}.

Uralic languages have received relatively moderate interest from the language
modeling community. Aside from the three national languages, no other Uralic
language is supported by any of the multilingual models, nor does any have a
monolingual model. There are no Uralic languages among the 15 languages
of XNLI. \citet{Wu:2020a} do explore all 100 languages that mBERT supports but
do not go into monolingual details. \citet{Alnajjar:2021} transfer existing
BERT models to minority Uralic languages, the only work that focuses solely on
Uralic languages.

In this paper we evaluate multilingual and monolingual models on Uralic
languages. We consider three evaluation tasks: morphological probing, POS
tagging and NER. We also use the models in a crosslingual setting, in other
words, we test how monolingual models perform on related languages.
We show that

\begin{itemize}
    \item these language models are very good at all three tasks when finetuned
        on a small amount of task specific data,
  \item for morphological tasks, when native BERT models are available (et,
    fi, hu), these outperform the others on their native language, though
    the advantage over XLM-RoBERTa is not statistically significant,
  \item for POS and NER, the use of native models from related, even closely
    related languages, rarely brings improvement over the multilingual
    models or even English models, 
\item as long as the alphabet that the language uses is covered in the
  vocabulary of the model, we can transfer mBERT (or RuBERT) to the NER and
  POS tasks with surprisingly little finetuning data.
\end{itemize}

\section{Approach}

We evaluate the models through three tasks: morphological probing, POS tagging
and NER.  Uralic languages have rich inflectional morphology and largely free
word order.  Morphology plays a key role in parsing sentences. Morphological
probing tries to recover morphological tags from the sentence representation
from these models.

For assessing the sentence level behavior of the models we chose two
token-level sentence tagging tasks, POS and NER. Part of speech tagging is a
common subtask of downstream NLP applications such as dependency parsing.
Named entity recognition is indispensable for various high level semantic
applications such as building knowledge graphs.  Our model architecture is
identical for POS and NER.

\subsection{Morphological probing}

\begin{figure}[t]
	\resizebox{0.49\textwidth}{!}{
		\begin{tikzpicture}
			\node (tokenizer) at (0, 0) [large_module] {subword tokenizer};
			\node (t1) at (-3, -\rowdist) [sequence] {You};
			\node (t2) at (-1.25, -\rowdist) [sequence] {have};
			\node (t3) at (1, -\rowdist) [target_sequence] {patience};
			\node (t4) at (3, -\rowdist) [sequence] {.};
			\foreach \x in {1,...,4}
			\draw [arrow,<-] (tokenizer.south) to[*|] (t\x.north);
			\node (wp1) at (-4.5, \rowdist) [sequence] {[CLS]};
			\node (wp2) at (-3, \rowdist) [sequence] {You};
			\node (wp3) at (-1.25, \rowdist) [sequence] {have};
			\node (wp4) at (0.25, \rowdist) [sequence] {pati};
			\node (wp5) at (1.5, \rowdist) [target_sequence] {\#\#ence};
			\node (wp6) at (3, \rowdist) [sequence] {.};
			\node (wp7) at (4.5, \rowdist) [sequence] {[SEP]};
			\foreach \x in {2,...,6}
			\draw [arrow] (tokenizer.north) to[*|] (wp\x.south);
			\node[fit=(wp4)(wp5)](wp_full){};

			\node (bert) at (0, 2*\rowdist) [bert_module] {contextualized model};
			\foreach \x in {1,...,7}
			\draw [arrow,<-] (bert.south) to[*|] (wp\x.north);

			\foreach \x in {-4.5,-3,-1.25,0.25,3,4.5} \drawbertoutput{\x}{3*\rowdist} ;
			\drawberttargetoutput{1.5}{3*\rowdist} ;
			\foreach \x in {-4.5,-3,-1.25,0.25,1.5,3,4.5}
			\draw [arrow] (bert.north) to[*|] (\x, 3*\rowdist-\matrixheight/2);

			\node (weight) at (1.5, 4*\rowdist) [small_module] {$\sum w_i x_i$};
			\node (mlp) at (-1, 4*\rowdist) [small_module] {MLP};
			\draw [arrow,<-] (weight.south) -- (1.5, 3*\rowdist+\matrixheight/2);
			\draw [arrow] (weight.west) -- (mlp.east);
			\node (output) at (-3.5, 4*\rowdist) [output] {$P(\text{label})$};
			\draw [arrow] (mlp.west) -- (output.east);


		\end{tikzpicture}
	}

	\caption{Probing architecture. Input is tokenized into subwords and a
		weighted average of the mBERT layers taken on the last subword of
		the target word is used for classification by an MLP. Only the MLP
		parameters and the layer weights $w_i$ are trained.}
	\label{fig:tikz_architecture}
\end{figure}
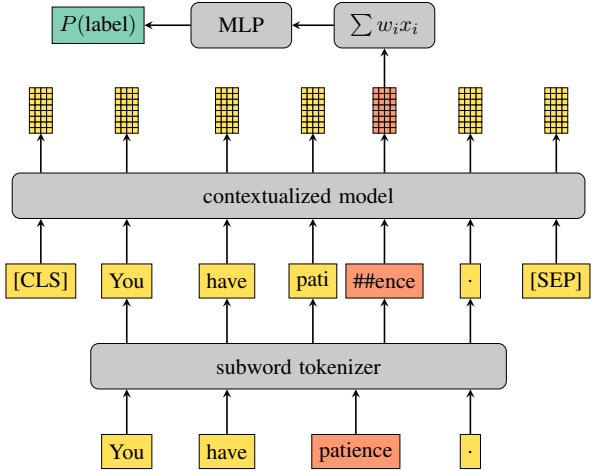

Probing is a popular evaluation method for black box models. Our approach is
illustrated in Figure~\ref{fig:tikz_architecture}.  The input of a probing
classifier is a sentence and a target position (a token in the sentence).  We
feed the sentence to the contextualized model and extract the representation
corresponding to the target token. Early experiments showed that lower layers
retain more morphological information than higher layers so instead of using
the top layer, we take the weighted average of all Transformer layers and the
embedding layer. The layer weights are learned along with the other parameters
of the neural network.  We train a small classifier on top of this
representation that predicts a morphological tag. We expose the classifier to a
limited amount of training data (2000 training and 200 validation instances).
If the classifier performs well on unseen data, we conclude that the
representation includes the relevant morphological information.

We generate the probing data for Estonian and Finnish from the Universal
Dependencies (UD) Treebanks
\citep{Nivre:2020,Haverinen:2013,Pyysalo:2015,Vincze:2010} and from the automatically
tagged Web\-corpus 2.0 for Hungarian since the Hungarian UD is very small.
Unfortunately we could not extend the list of languages to other Uralic
languages because their treebanks are too small to sample enough data.

The sampling method is constrained so that the target words have no overlap
between train, validation and test, and we limit class imbalance to 3-to-1
which resulted in filtering some rare values. We were able to generate enough
probing data for 11 Estonian, 16 Finnish and 11 Hungarian tasks, see 
Table~\ref{table:morph_tasks} for the full list of these.

\subsection{Sequence tagging tasks}

Our setup for the two sequence tagging tasks is similar to that of the
morphological probes except we train a shared classifier on top of all token
representations. We use the vector corresponding to the first subword in both
tasks. Although this may be suboptimal in morphology, \citet{Acs:2021b} showed
that the difference is smaller for POS and NER. We also finetune the models
which seems to close the gap between first and last subword pooling for
morphology, see \ref{ss:morph}. For sequence tagging tasks, unlike for
morphology, we found that the weighted average of all layers is suboptimal
compared to simply using the top layer, so the experiments presented here all
use the top layer.

We sample 2000 train, 200 validation and 200 test sentences as POS training
data from the largest UD treebank in Estonian and Finnish, and from Webcorpus
2.0 for Hungarian. Aside from these three, Erzya [myv]; Moksha [mdf]; Karelian
[krl]; Livvi [olo]; Komi Permyak [koi]; Komi Zyrian [kpv]; Northern Sámi [sme];
and Skolt Sámi [sms] have UD treebanks
\citep{Rueter:2018,Rueter:2018b,Pirinen:2019,Rueter:2014,Rueter:2020,Partanen:2018,Sheyanova:2017},
but these are considerably smaller in size.  Although none of these languages
are officially supported by any of the language models we evaluate, we train
crosslingual models and find that the models have remarkable crosslingual
capabilities.

Our NER data is sampled from WikiAnn \citep{Pan:2017}. WikiAnn has data in
Erzya, Estonian, Finnish, Hungarian, Komi Permyak, Komi Zyrian, Moksha,
and Northern Sámi.\footnote{WikiAnn also has Udmurt data, but the transcription is
problematic: Latin and Cyrillic are used inconsistently, Wikipedia Markup is
parsed incorrectly etc.} Similarly to the POS training data,
we sample 2000 training, 200 validation and 200 test sentences when available,
see Table~\ref{tab:training_size} for actual training set sizes. 

\begin{table}
	\centering
    \resizebox{0.48\textwidth}{!}{
	\begin{tabular}{l l r r r}
		\toprule
		Language & Code & Morph & POS & NER \\
		\midrule
		Hungarian & {[hu]} &26k&2000& 2000 \\
		Finnish & {[fi]} &38k&2000& 2000\\
		Estonian & {[et]} &26k&2000& 2000\\
		Erzya & {[myv]} &0&1680& 1800\\
		Moksha & {[mdf]} &0&164& 400\\
		Karelian & {[krl]} &0&224& 0\\
		Livvi & {[olo]} &0&122& 0\\
		Komi Permyak & {[koi]} &0&78& 2000\\
		Komi Zyrian & {[kpv]} &0&562& 1700\\
		Northern Sámi & {[sme]} &0&2000& 1200\\
		Skolt Sámi & {[sms]} &0&101& 0\\
		\bottomrule
	\end{tabular}
}
    \caption{\label{tab:training_size} Size of training data for each language.}
\end{table}

\subsection{Training details}

We train all classifiers with identical hyperparameters. The classifiers have
one hidden layer with 50 neurons and ReLU activation. The input and the output
dimensions are determined by the choice of language model and the number of
target labels. The classifiers have 40 to 60k trainable parameters which are
randomly initialized and updated using the backpropagation algorithm.  We run
experiments both with and without finetuning the language models. Finetuning
involves updating both the language model (all 110M parameters) and the
classification layer (end-to-end training).

All models are trained using the AdamW optimizer \citep{Loshchilov:2019} with
$lr=0.0001, \beta_1 = 0.9, \beta_2 = 0.999$. We use 0.2 dropout for
regularization and early stopping based on the development set.  We set the
batch size to 128 when not finetuning the models, and we use batch size 8, 12 or
20 when we finetune them.

The evaluated models, all from the BERT/RoBERTa family, differ only in the
choice of training data and the training objective. They all have 12
Transformer layers, with 12 heads, and 768 hidden dimensions, for a total of
110M parameters.

\section{The models evaluated}
\label{sec:models}

Our goal is twofold: we want to assess monolingual models against
multilingual models, and we want to evaluate the models on 'unsupported'
languages, both typologically related and unrelated. 

We pick two multilingual models, mBERT and XLM-RoBERTa. Our choices for
monolingual models are EstBERT for Estonian, FinBERT for Finnish and
HuBERT for Hungarian (See Table~\ref{table:model_list}). As a control, we also
test the English BERT as a general test for cross-language transfer. Since
many Uralic speaking communities are in Russia and the languages are heavily
influenced by Russian, we test RuBERT on these languages. Finally, we also
test a randomly initialized mBERT.  We do this because the capacity of the
BERT-base models is so large that they may memorize the probing data alone.
Many models have cased and uncased version, the latter often removing
diacritics along with lowercasing. Since diacritics play an important role in
many Uralic languages, we only use the cased models.  We return to this issue
in \ref{sec:segmentation}.

The models along with their string identifier are summarized in
Table~\ref{table:model_list}.

\begin{table*}
    \centering
    \resizebox{\textwidth}{!}{
    \begin{tabular}{llll}
        \toprule
        Model & Identifier & Language(s) &  Reference \\
        \midrule
        mBERT & bert-base-multilingual-cased & 100+ inc. et, fi, hu & \citet{Devlin:2018a} \\
        XLM-RoBERTa & xlm-roberta-base & 100 inc. et, fi, hu & \citet{Liu:2019a} \\
        EstBERT & tartuNLP/EstBERT & Estonian & \citet{Tanvir:2021} \\
        FinBERT & TurkuNLP/bert-base-finnish-cased-v1 & Finnish & \citet{Virtanen:2019}\\
        HuBERT & SZTAKI-HLT/hubert-base-cc & Hungarian & \citet{Nemeskey:2020} \\
        EngBERT & bert-base-cased & English & \citet{Devlin:2018a} \\
        RuBERT & DeepPavlov/rubert-base-cased & Russian & \citet{Kuratov:2019} \\
        rand-mBERT & mBERT with random weights & any & described in Section~\ref{sec:models} \\
        \bottomrule
    \end{tabular}}
    \caption{\label{table:model_list} List of models we evaluate.}
\end{table*}

\subsection{Subword tokenization}\label{sec:segmentation}

Subword tokenization is a key component in achieving good performance on
morphologically rich languages.  There are two different tokenization methods
used in the models we compare: XLM-RoBERTa uses the SentencePiece algorithm
\cite{Kudo:2018b}, the other models use the WordPiece algorithm
\cite{Schuster:2012}.  The two types of tokenizers are algorithmically very
similar, the differences between them are mainly dependent on the vocabulary
size per language.  The multilingual models consist of about 100 languages,
and the vocabularies per language apper sublinearly proportional to the amount
of training data available per language: in case of  mBERT, 77\% of the
word pieces are pure ascii \citep{Acs:2019d}.

The native models, trained on monolingual data, have longer and more
meaningful subwords (see the bolded entries in Table~\ref{table:subword}).
This greatly facilitates the sharing of train data, a matter of great
importance for Uralic languages where there is little text available to begin
with.

Both BERT- and RoBERTa-based models first tokenize along whitespaces, but the
handling of missing characters differs significantly. In BERT-based models, if
there is a character missing from the tokenizer's vocabulary, the model
discards the whole segment between whitespaces, labeling it [UNK].  In
cross-lingual cases many words are lost since monolingual models tend to lack
the extra characters of a different language. In contrast, XLM-RoBERTa deletes
the unknown characters, but the string that remains between whitespaces is
segmented, so the loss of information is not as severe.

\begin{table*}
	\centering
	\resizebox{\textwidth}{!}{
	\begin{tabular}{l r r r r r r r}
		\toprule
		         & mBERT & RoBERTa & EstBERT & FinBERT & HuBERT & RuBERT & EngBERT\\
		\midrule
		Vocab. size & 120k & 250k & 50k & 50k & 32k & 120k & 29k\\
		\midrule
		Missing  [et] (\%)& .0 & .0 &   \textbf{.2} &   .0 & .5 &  .1 & .2\\
		Missing  [fi] (\%)& .0  & .0  &   .0 &   \textbf{.0} & .4 &  .0 & .0\\
		Missing  [hu] (\%) & .1 & .0 & 21.5 & 48.3 & \textbf{.1} & 2.7 & .2\\
		Missing  [sme] (\%)& .2 & .0 & 15.0 & 47.4 & 5.1& 4.8 & .2\\
		Missing  [myv] (\%)& .0 & .0 & 97.5 & 97.5 &97.5&  .0 & .0\\
		\midrule
		Subword length [et] & 3.7$\pm$1.4 & 4.2$\pm$1.7 & \textbf{5.8}$\pm$2.6 & 3.7$\pm$1.4 & 3.1$\pm$1.2 & 3.1$\pm$1.2 & 3.5$\pm$1.4\\
		Subword length [fi]  & 3.8$\pm$1.4 & 4.5$\pm$1.9 & 3.8$\pm$1.4 & \textbf{5.9}$\pm$2.5 & 3.1$\pm$1.1 & 3.1$\pm$1.1 & 3.4$\pm$1.4\\
		Subword length [hu]& 3.5$\pm$1.5 & 4.2$\pm$2.0 & 3.3$\pm$1.2 & 3.1$\pm$1.1 & \textbf{5.0}$\pm$2.4 & 3.0$\pm$1.1 & 3.3$\pm$1.4\\
		Subword length [sme] &3.2$\pm$1.0&3.4$\pm$1.1& 3.2$\pm$1.1 & 3.2$\pm$1.1 & 3.1$\pm$1.2 & 2.9$\pm$1.0 & 3.0$\pm$1.0\\
		Subword length [myv] & 3.1$\pm$1.2 & 3.2$\pm$1.0 & 1.0$\pm$0.0 & 1.0$\pm$0.0 & 1.0$\pm$0.0 & 3.4$\pm$1.2 & 1.1$\pm$0.4\\
		\midrule
		Character length [et] & 9.2 & 9.2 &   \textbf{9.2} &   9.2 & 9.2 &  9.2 & 9.2\\
		Character length [fi]  & 9.3  & 9.3  & 9.3 & \textbf{9.3} & 9.3 & 9.3 & 9.3\\
		Character length [hu]& 9.8  & 9.8 &  9.6 & 8.8 & \textbf{9.8} & 9.8 & 9.9\\
		Character length [sme] & 8.5 & 8.5 & 8.3 & 7.6 & 8.5 & 8.4 & 8.5\\
		Character length [myv]    & 7.3 & 7.3 & 1.8 & 1.8 &1.7&  7.3 & 7.3\\
		\midrule
		Fertility [et] & 3.4 & 2.8 &   \textbf{2.1} &   3.6 & 4.4 &  4.3 & 4.3\\
		Fertility [fi]  & 3.3  & 2.7  & 3.5 & \textbf{1.9} & 4.6 & 4.4 & 4.5\\
		Fertility [hu]& 4.0  & 3.2 &  5.2 & 4.5 & \textbf{2.8} &  5.4 & 5.6\\
		Fertility [sme] & 3.7 & 3.6 & 4.1 & 3.3 & 4.5& 4.6 & 4.7\\
		Fertility [myv]    & 3.6 & 3.3 & 1.1 & 1.1 &1.1&  3.0 & 7.2\\
		\bottomrule
	\end{tabular}}
    \caption{\label{table:subword} Major characteristics of cross-language tokenization. Boldface font marks the corresponding language-model pairs.}
\end{table*}

Table~\ref{table:subword} summarizes different measures in language-model
pairs. As a general observation, Latin script models (FinBERT, HuBERT,
EstBERT) are unusable on Cyrillic text, as seen e.g.~on Erzya, where Latin
script models produce [UNK] token for 97.5\% of the word types. This is also
seen for Northern Sámi and Hungarian, which have many non-ascii characters (á,
é, í, ó, ö, ő, ú, ü, ű for Hungarian, č, đ, \textipa{\ng}, š, ŧ, ž for Northern Sámi)
see the Hungarian-EstBert/FinBERT pairs and the Northern Sámi-FinBERT/HuBERT
pairs.

The mean subword length generally lies between 3.0 and 3.5 for most pairs -
naturally, the corresponding language-model pairs have much higher mean
subword length, 5.0 to even 5.9.  This range is true not only for Latin script
languages, but for Cyrillic script languages as well, as indicated by Erzya,
which has a mean subword length of 3.1 to 3.4 on the multilingual models and
on RuBERT.

Fertility \citep{Acs:2019d} is defined 
as the average number of BERT word pieces found in a single real word type.
EstBERT on Estonian and FinBERT on Finnish have very similar fertility values
(2.1 and 1.9), but HuBERT on Hungarian has much higher fertility. This is
mainly caused by the different vocabulary sizes - the Finnic models have 50000
subwords in their vocabulary, HuBERT only contains 32000 subwords.  The rest
of the fertility values are mostly over 3. In extreme cases, a word is
segmented into letters, which is the case for EngBERT on Erzya, but the
non-Hungarian models on Hungarian also produce very high fertility values.

\begin{figure*}[ht]
    \centering
    \includegraphics[width=\textwidth]{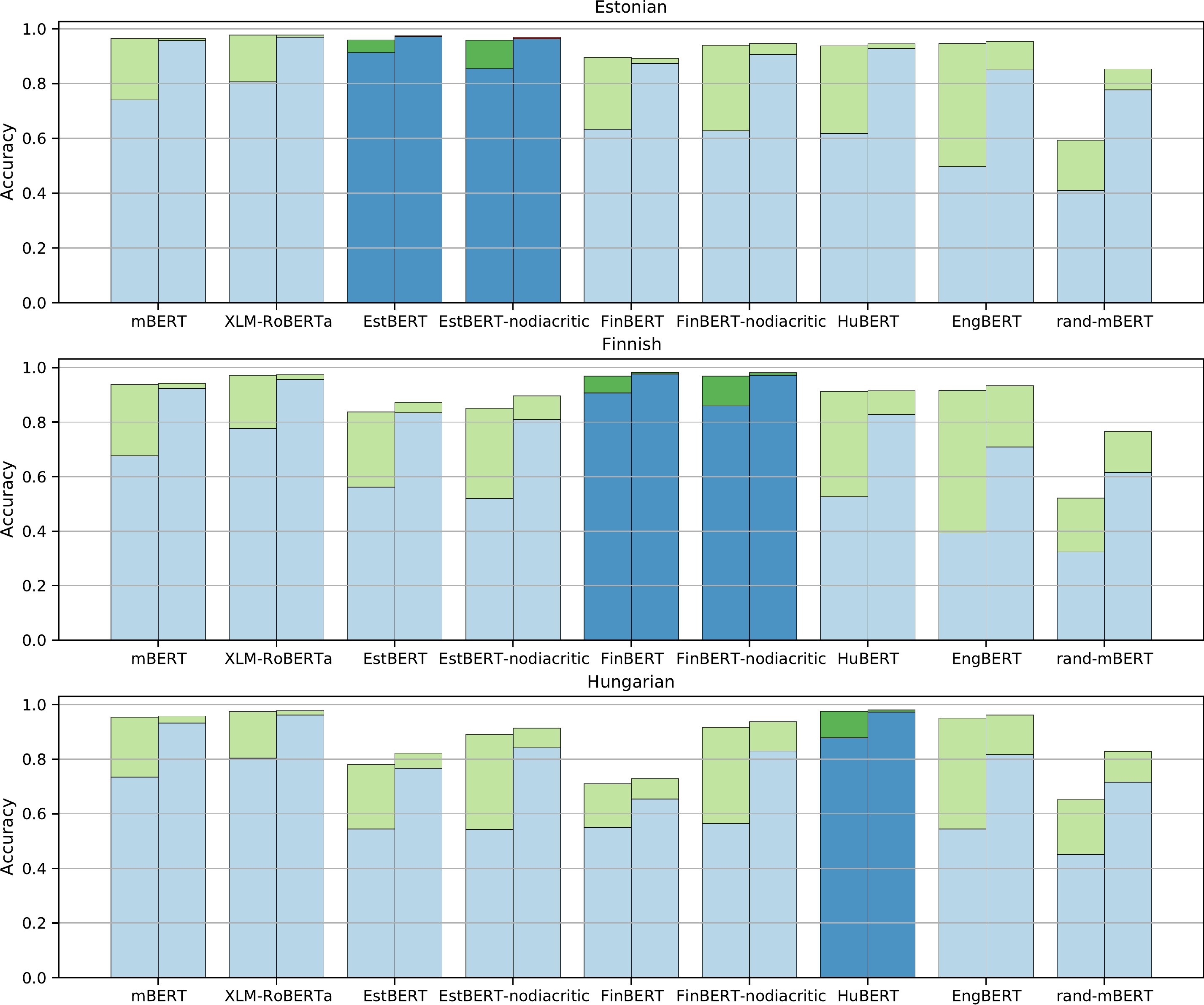}
    \caption{Mean accuracy of morphological tasks by language. The bars are
    grouped in two, the left one is the result of probing the first subword,
the right one is the results of probing the last subword. Blue bars are without
finetuning, green bars are with finetuning. Monolingual models are
highlighted.}
    \label{fig:morph_first_last}
\end{figure*}

\section{Results}

\subsection{Morphology}\label{ss:morph}

Morphological tasks are generally easy for most models and we see reasonable
accuracy from crosslingual models as illustrated by
Figure~\ref{fig:morph_first_last}. Mean accuracies, especially after
finetuning, are generally above 90\%, except, unsurprisingly, for the randomly
initialized models.

\paragraph{Subword choice} We first start by examining the choice of subword
on morphological tasks. We try probing the first and the last subword and we
find that there is a substantial gap in favor of the last subword. This is
unsurprising considering that Uralic languages are mainly suffixing. This gap
on average shrinks from 0.21 to 0.032 when we finetune the models on the
probing data (Figure~\ref{fig:morph_first_last} shows this gap in green).
Without finetuning there is only one task, \task{Hungarian}{Degree}{ADJ}, where
probing the first subword is better than probing the last one for some models.
This is explained by the fact that the superlative in Hungarian is formed from
the comparative by a prefix.

\paragraph{Monolingual models} are only slightly better than the two
multilingual models, XLM-RoBERTa in particular. We run paired t-tests on the
accuracy of each model pair over the 11 (et, hu) or 16 (fi) morphological tasks
in a particular language and find that the difference between the
monolingual model and XLM-RoBERTa is never significant, and for Estonian,
neither is the difference between EstBERT and mBERT.

\paragraph{Cross-lingual transfer} works only if we finetune the models.
Interestingly, language relatedness does not seem to play a role here. FinBERT
transfers worse to Estonian than HuBERT, and EstBERT transfers worse to Finnish than
HuBERT. Interestingly, EngBERT transfers better to all three models than the
other native BERTs, and for Finnish and Hungarian it is actually on par with
mBERT.

\paragraph{Diacritics} As seen from the first panel of
Table~\ref{table:subword}, EstBERT and FinBERT replace words with unknown
characters with [UNK] to such an extent that a large proportion of types end up
being filtered. We try to mitigate this issue by preemptively removing all
diacritics from the input.  It appears that this has little effect on the
original language, but cross-lingual transfer is improved for Finnish. In the
sequence tagging tasks that we now turn to, we remove the diacritics when we
evaluate EstBERT or FinBERT in a cross-lingual setting.

\subsection{POS and NER}

\begin{figure}[ht]
    \centering
    \includegraphics[width=0.5\textwidth]{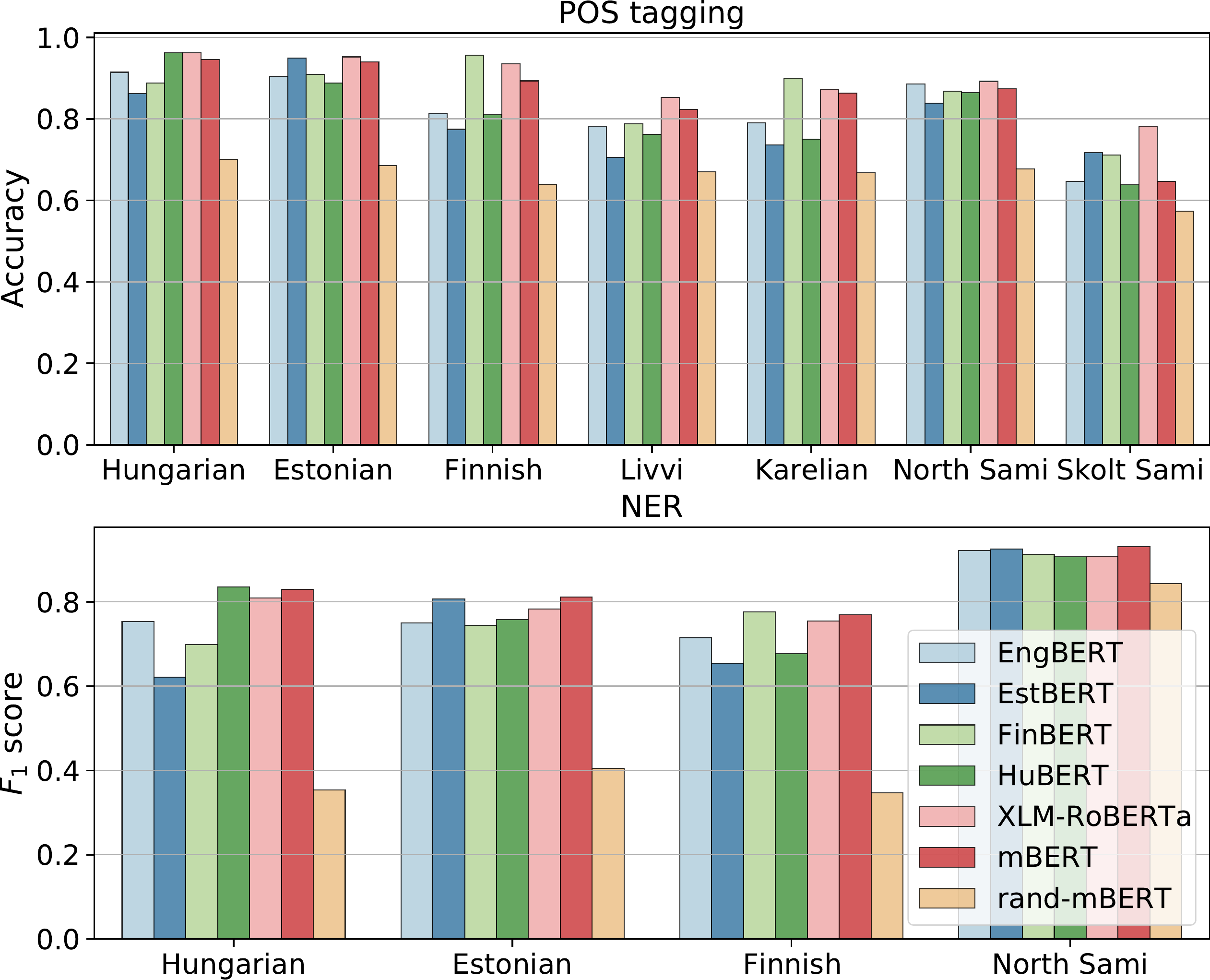}
    \caption{POS and NER results on languages that use the Latin
    alphabet.}
    \label{fig:pos_ner_latin}
\end{figure}

\begin{figure}[ht]
    \centering
    \includegraphics[width=0.49\textwidth]{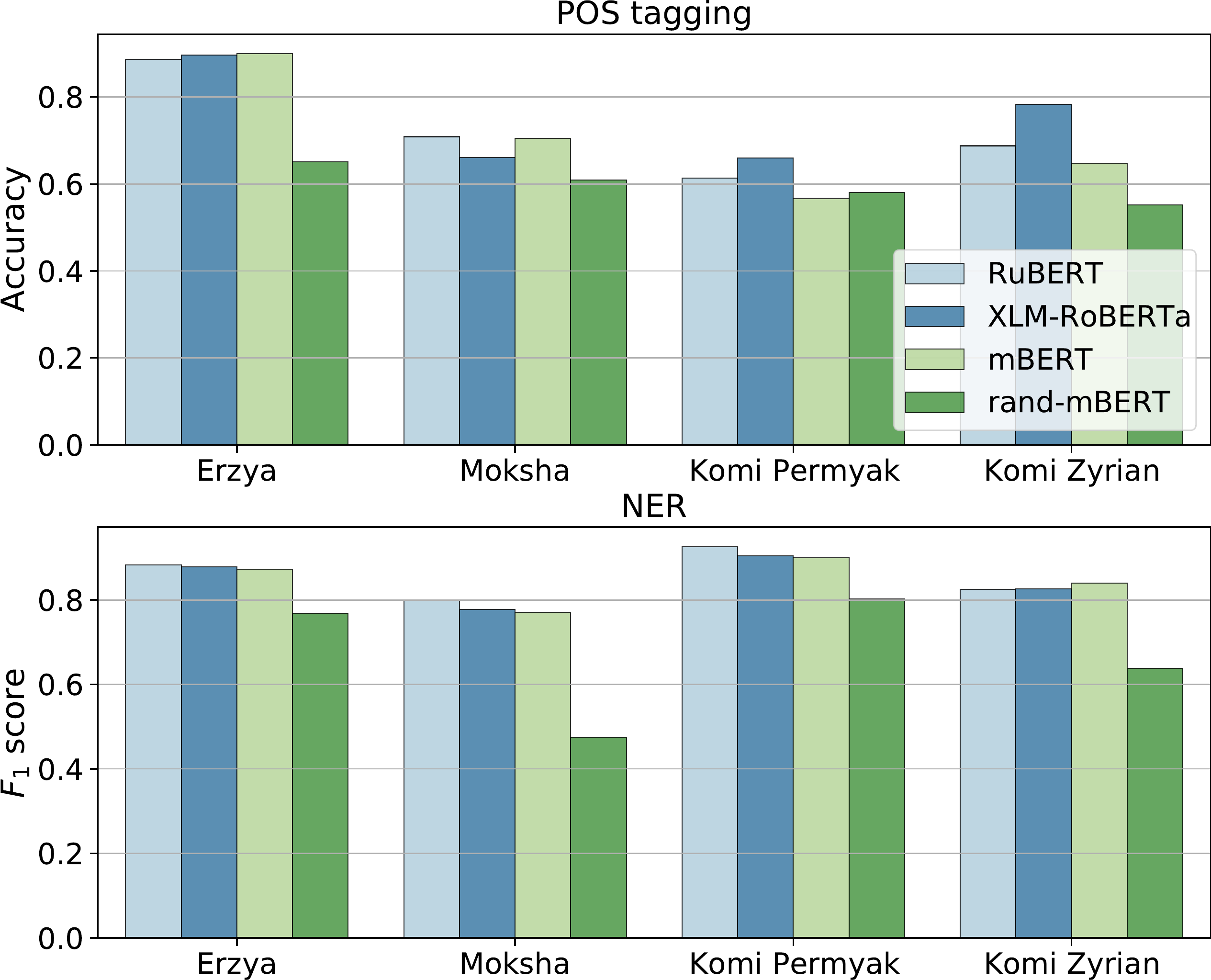}
    \caption{POS and NER results on languages that use the Cyrillic alphabet.}
    \label{fig:pos_ner_cyrillic}
\end{figure}

We extend our studies to all Uralic languages with any training data (see
Table~\ref{tab:training_size}) and we limit the discussion to finetuned models
since cross-lingual transfer does not work without finetuning.
We split the languages into two groups, Latin and Cyrillic, and we only test
models with explicit support for the script that the language uses.
Multilingual models support both scripts.
Figures~\ref{fig:pos_ner_latin} and \ref{fig:pos_ner_cyrillic} show the
results by language.

\paragraph{National languages} We generally find the best performance in the
three languages with native support: Estonian, Finnish and
Hungarian. Monolingual models perform the best in their respective
language but the two multilingual models are also very capable.

\paragraph{Cross-lingual transfer} does not seem to benefit
from language relatedness, EngBERT transfers just as well as other monolingual
models. Even extremely close relatives such as Livvi and Finnish do
not transfer better than XLM-RoBERTa to Livvi. On the other hand, FinBERT is
the best for Karelian POS, another close relative of Finnish. The writing
system and shared vocabulary also seem to play an important role, as seen from
RuBERT's usefulness on unrelated but Cyrillic-using Uralic languages,
see Figure~\ref{fig:pos_ner_cyrillic}.

\paragraph{XLM-RoBERTa} is generally a strong model for cross-lingual transfer
for all Uralic languages. We suspect that this is due to its large subword
vocabulary, which may provide a better generalization basis for capturing the
orthographic cues that are often highly indicative in agglutinative languages.

\paragraph{North Sámi} Both POS and NER in North Sámi are relatively easy as
long as the orthographic cues can be captured (i.e.~the Latin script is
supported). rand-mBERT is suprisingly successful at NER in North Sámi,
suggesting that orthograpic cues (rand-mBERT uses mBERT's tokenizer) are
highly predictive of named entities in North Sámi.


\begin{table*}[t]
	\centering
	\begin{tabular}{lllll}
		\toprule
		Morph tag & POS & Estonian & Finnish & Hungarian\\
		\midrule
        Case & adj & 8 classes& 11 classes & \\
        Case & noun & 15 classes & 12 classes & 18 classes \\
        Case & propn & & 8 classes & \\
        Case & verb & & 12 classes & \\
        \midrule
        Degree & adj & & Cmp, Pos, Sup & Cmp, Pos, Sup \\
        \midrule
        Derivation & adj & & Inen, Lainen, Llinen, Ton & \\
        Derivation & noun & & Ja, Lainen, Minen, U, Vs & \\
        \midrule
        InfForm & verb & & 1, 2, 3 & \\
        \midrule
        Mood & verb & & Cnd, Imp, Ind & Cnd, Imp, Ind, Pot\\
        \midrule
		Number psor & noun & & & Sing, Plur \\
        Number & a/n/v & Sing, Plur & Sing, Plur & Sing, Plur \\
        \midrule
        PartForm & verb & & Pres, Past, Agt & \\
        \midrule
		Person psor  & noun & & & 1, 2, 3\\
        Person & verb & 1, 2, 3 & & 1, 2, 3 \\
        \midrule
		Tense & adj & Pres, Past && \\
		Tense & verb & Pres, Past & Pres, Past & Pres, Past \\
        \midrule
        VerbForm & verb & Conv, Fin, Inf, Part, Sup & Inf, Fin, Part & Inf, Fin \\
        \midrule
        Voice & adj & Act, Pass & & \\
        Voice & verb & Act, Pass & Act, Pass & \\
		\bottomrule
	\end{tabular}
	\caption{\label{table:morph_tasks} List of morphological probing tasks.}
\end{table*}

\section{Conclusion}\label{sec:conc}

Altogether we find that it is possible, and relatively easy, to transfer
models to new languages with finetuning on very limited training data, though
extremely limited data still hinders progress: compare Erzya (1680 train
sentences) to Moksha (164 train sentences) on Fig.~\ref{fig:pos_ner_cyrillic}.

EngBERT and RuBERT, which we introduced as a control for language transfer
among genetically unrelated languages, transfer quite well: in particular
the Latin-script EngBERT transfers better to Hungarian than FinBERT or
EstBERT.

We note that we did not perform monolingual hyperparameter search or any
preprocessing, and there is probably room for improvement for each of these
languages. The biggest immediate gains are expected from extending the UD and
WikiAnn datasets, and from careful handling of low-level characterset and
subword tokenization issues. There are many Uralic languages that still lack
basic resources, in particular the entire Samoyedic branch, Mari, and Ob-Ugric
languages, are currently out of scope.  Another avenue of research could be to
work towards a stronger mBERT interlingua, or perhaps one for each script
family, as the charset issues are clearly relevant.

Our data, code and the full result tables are available at \url{https://github.com/juditacs/uralic_eval}.

\section*{Acknowledgements}
This work was partially supported by the Ministry of Innovation and the
National Research, Development and Innovation Office within the framework of
the Artificial Intelligence National Laboratory Programme.


\bibliographystyle{acl_natbib}
\bibliography{ml}

%
\end{document}